\documentclass[letterpaper, 10 pt, conference]{ieeeconf}  %

\IEEEoverridecommandlockouts                              %

\usepackage{graphicx} %
\usepackage{times}
\usepackage{amsmath,amssymb}  %
\usepackage{verbatim}
\usepackage{subfigure}
\usepackage{mathtools}
\usepackage{algpseudocode}
\usepackage[boxed]{algorithm2e}
\usepackage{tabularx} 
\usepackage{multirow}
\usepackage{soul, color} 
\usepackage{float}
\usepackage{caption}
\usepackage[table,x11names]{xcolor}
\usepackage{url}
\usepackage{hyperref}
\usepackage{cite}

\usepackage[export]{adjustbox}
\usepackage{rotating}
\usepackage{booktabs, makecell, tabularx}
\let\labelindent\relax
\usepackage{enumitem}

\usepackage{xcolor}

\SetKwRepeat{Do}{do}{while}

\makeatletter
\def\endthebibliography{%
  \def\@noitemerr{\@latex@warning{Empty `thebibliography' environment}}%
  \endlist
}
\makeatother

\title{\LARGE \bf
\centering Localization with Anticipation for Autonomous Urban Driving in Rain
}

\author{Yu Xiang Tan, Malika Meghjani and Marcel Bartholomeus Prasetyo
     \thanks{Yu Xiang Tan, Malika Meghjani and Marcel Bartholomeus Prasetyo are with Singapore University of Technology and Design, Singapore. {\tt\small yuxiang\_tan@mymail.sutd.edu.sg, \{malika\_meghjani, marcel\_prasetyo\}@sutd.edu.sg}} %
}

\begin{document}

\onecolumn %
\pagestyle{empty} %
\begin{center}
  \large\bfseries  %
  This work has been submitted to the IEEE for possible publication. Copyright may be transferred without notice, after which this version may no longer be accessible.
\end{center}
\twocolumn %

\maketitle
\thispagestyle{empty}
\pagestyle{empty}

\begin{abstract}
This paper presents a localization algorithm for autonomous urban vehicles under rain weather conditions. In adverse weather, human drivers anticipate the location of the ego-vehicle based on the control inputs they provide and surrounding road contextual information. Similarly, in our approach for localization in rain weather, we use visual data, along with a global reference path and vehicle motion model for anticipating and better estimating the pose of the ego-vehicle in each frame. The global reference path contains useful road contextual information such as the angle of turn which can be potentially used to improve the localization accuracy especially when sensors are compromised. We experimented on the Oxford Robotcar Dataset and our internal dataset from Singapore to validate our localization algorithm in both clear and rain weather conditions. Our method improves localization accuracy by $50.83$\% in rain weather and $34.32$\% in clear weather when compared to baseline algorithms. 

\end{abstract}

\section{Introduction}

Robust localization is required for autonomous vehicles to operate in all weather conditions. Recent approaches use radar sensors as the primary input into their localization algorithms \cite{hong_radarslam_2020, barnes_under_2020, ort_autonomous_2020} as they are less affected by adverse weather conditions compared to cameras. However, these sensors can be significantly more expensive and computationally costly. 
In this work, we aim to use visual data for localization using stereo camera which is relatively a cheaper sensor and computationally less expensive.

\begin{figure}[t]
 \centering
    \includegraphics[width=0.85\linewidth]{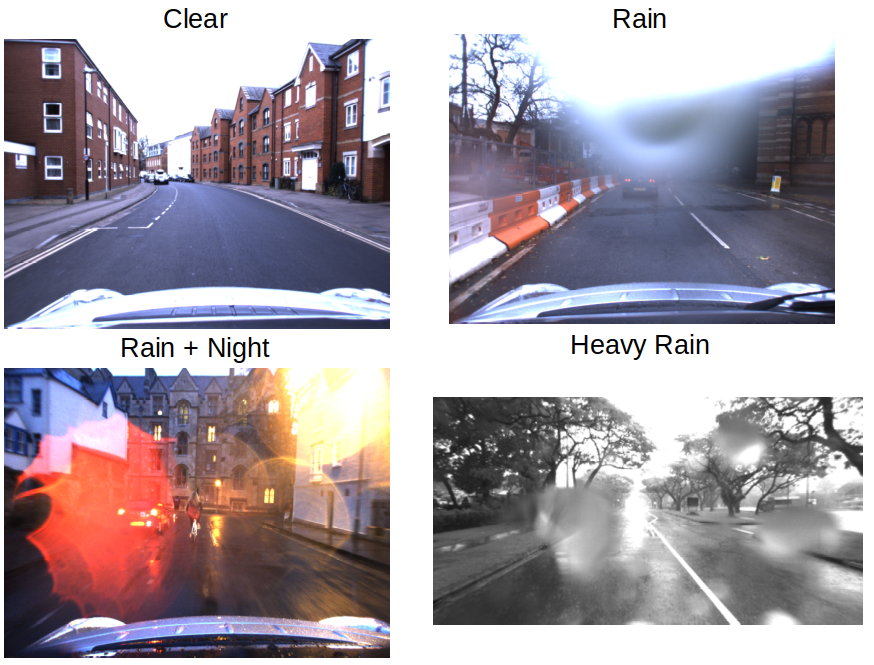}
    \caption{Various rain conditions from the Oxford Robotcar Dataset \cite{maddern_1_2017} (top-left, top-right, bottom-left) and Singapore Dataset (bottom-right) }
\label{fig:various_rain_conditions}
\end{figure}

In rain weather, visual data is easily compromised due to noise artefacts such as lens-flare under bright lighting, raindrops on the camera lenses or a combination of both, similar to the examples illustrated in Fig. \ref{fig:various_rain_conditions}. As such, visual features become less reliable which in turn reduces the accuracy of visual localization algorithms. We tackle this problem by introducing additional information from the vehicle's global reference path and its motion model to complement the compromised visual data.

The global reference path (GRP) is a route for the vehicle to follow, manually generated by the user using a routing service. This is similar to how drivers use navigation systems to search for a route to an unfamiliar location before driving. Assuming that the vehicle follows the route provided, the GRP could serve as either an initial estimate for visual odometry algorithms or a standalone source of localization.

We build our proposed method on top of DROID-SLAM \cite{teed_droid-slam_2021}, a state-of-the-art visual SLAM algorithm, and validate it on both ideal and rain weather data from two cities. We compare these localization results with ORB-SLAM3 \cite{campos_orb-slam3_2021} and DROID-SLAM. 

Our proposed approach improved the localization accuracy over the DROID-SLAM algorithm by $50.75$\% in rain weather and $34.32$\% in clear weather. We validated the use of the GRP for improving localization across a range of rain weather conditions.

The contributions of our work are:
\begin{enumerate}[label=(\alph*)]
    \item An improved and reliable pose estimate of the ego-vehicle by using the GRP
    \item A robust set of heuristics which dynamically controls the keyframe selection frequency based on the confidence of feature correspondences
    \item Validation on real-world urban driving data from two climatically diverse cities (Oxford and Singapore) and across a range of rain weather conditions which are compared against the clear weather localization results.
\end{enumerate}

\section{Related Work}

\subsection{Localization in Adverse Weather}

Performing localization in adverse weather is a challenging problem as visual features can appear differently in different weathers. Specifically in rain, there is an occlusion of visual features due to an overexposure of the image or distortions caused by raindrops on the lenses. The current approaches to deal with this problem include: (a) Using sensors, such as radar, which are robust to noise  \cite{hong_radarslam_2020, ort_autonomous_2020}. For example, Radar SLAM from \cite{hong_radarslam_2020} outperforms vision and LiDAR based algorithms for localization in adverse weather conditions. However, radar sensors can be significantly
more expensive and computationally costly. (b) Removing adverse weather effects from sensor information to improve localization accuracy \cite{yu_visual_2021, porav_adversarial_2018}. Porav et. al. \cite{porav_adversarial_2018} designed an image transforming filter to transform an adverse weather scene into an ideal weather scene. 
This helps improve feature matching and thus localization accuracy. However, this method requires paired image data from adverse weather and ideal weather for the same route which might not always be feasible.
(c) Generating robust visual feature detectors and descriptors that are more invariant to noise \cite{revaud_r2d2_2019, dusmanu_d2-net_2019, sarlin_back_2021}. R2D2 \cite{revaud_r2d2_2019} and D2-Net \cite{dusmanu_d2-net_2019} uses deep learning to train their feature detectors and descriptors to be more robust even in challenging conditions. Although these methods were not directly evaluated on adverse weather localization, they show improved feature matching under illumination and viewpoint changes.
These approaches could be further enhanced by introducing additional sources of information. In our approach, we augment the visual data with the vehicle's GRP and motion model information.
This introduces a fourth category of approach to address the localization problem in adverse weather, where all four approaches are not mutually exclusive.

\subsection{Map-based Localization}
Brubaker et. al. uses visual odometry and road map data to perform self-localization \cite{brubaker_lost_2013}. They assume the visual odometry is accurate and feeds the output into a filtering algorithm to figure out where the vehicle is within the city. Our work also uses both visual odometry and road map data. However, our focus is on localizing in rain weather where visual odometry is likely to be inaccurate. Thus, we adopt a different approach where the road map data is used to improve the accuracy of the visual odometry.
\cite{ma_exploiting_2019, guo_coarse--fine_2021} use High Definition Maps (HD Maps) and road lane marking to perform localization. The HD Map approaches aim to match road signs, observed from a camera on the vehicle, to a top down view of a previously-obtained map. This approach requires well-trained segmentation algorithms to identify and match the road signs, which might not perform reliably in rain conditions. In addition, the HD Map would also be difficult to obtain and large in size. Comparatively, our method builds upon the well explored field of visual odometry and only requires a small memory size to store the robot's global reference path.

\subsection{Visual Odometry}
Visual Odometry (VO) approaches can be mainly categorized as indirect or direct approaches. 
\subsubsection{Indirect approaches}
Indirect approaches process the input image by identifying keypoints using feature detectors and labelling them with feature descriptors \cite{teed_droid-slam_2021,mur-artal_orb-slam_2015,mur-artal_orb-slam2_2017,campos_orb-slam3_2021,klein_parallel_2007,geiger_stereoscan_2011}. These features are usually keypoints on the image, with its neighbourhood forming its descriptor. Thus, distortions and noise from adverse weather could negatively impact the keypoint detection and descriptor's accuracy and reliability.
Early works of indirect approaches include PTAM \cite{klein_parallel_2007} and VISO2 \cite{geiger_stereoscan_2011} which explore different handcrafted features to perform localization. More recent work includes ORB-SLAM3 \cite{campos_orb-slam3_2021} which uses ORB features \cite{rublee_orb_2011} that are widely applicable and resistant to noise, and DROID-SLAM \cite{teed_droid-slam_2021} which uses dense deep-learned features that are trained on synthetic all-weather drone data. Despite the work done in designing features that are more resistant to noise, localization accuracy still suffers, leading us to explore other sources of information such as the GRP and vehicle motion model.

\subsubsection{Direct Approaches}
Direct approaches solve for the camera pose by minimizing the photometric error \cite{fleet_lsd-slam_2014, engel_direct_2016}. This method allows for more robust localization in low textured scenes and is also more robust in the presence of photometric noise. The drawback however, includes a higher computation cost and a more complex optimization problem to be solved. Additionally, the compression technique used in the Oxford Robotcar Dataset severely affects the accuracy of direct methods and thus this paper focuses the evaluation on indirect approaches.

\section{Approach} \label{approach}

\begin{figure}[h]
 \centering
    \includegraphics[width=\linewidth]{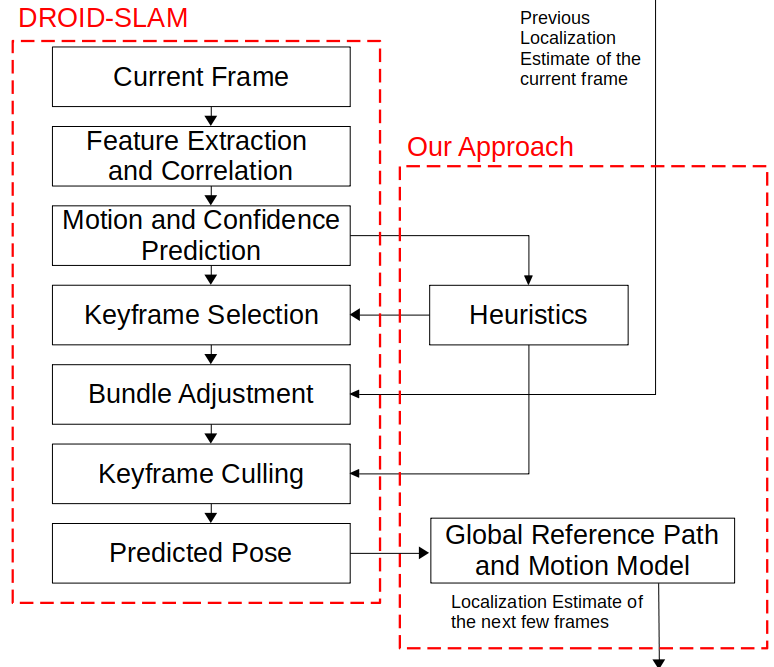}
    \caption{Integrated system overview of our approach complementing the DROID-SLAM algorithm}
\label{fig:overview}
\end{figure}

Our proposed approach for localization in rain is developed to complement any conventional visual odometry algorithm that uses keyframes. In our implementation, we
use DROID-SLAM, which is trained on synthetic all-weather conditions, as the base vision algorithm. We then add onto it additional map data while modifying their keyframe selection and culling strategies. Fig. \ref{fig:overview} gives a pictorial representation of how our approach complements DROID-SLAM. A localization estimate using the GRP (map data) targets the problem of increased localization errors caused by raindrops or lens flare while the heuristic methods of modifying keyframe selection and culling strategies targets the problem of tracking failures. Section \ref{localization_estimate} gives more details on how the GRP is used to perform localization estimation and Section \ref{heuristics} gives more details on the heuristic methods used. 

\subsection{Localization Estimate} \label{localization_estimate}
\begin{figure}[h]
 \centering
    \includegraphics[width=\linewidth]{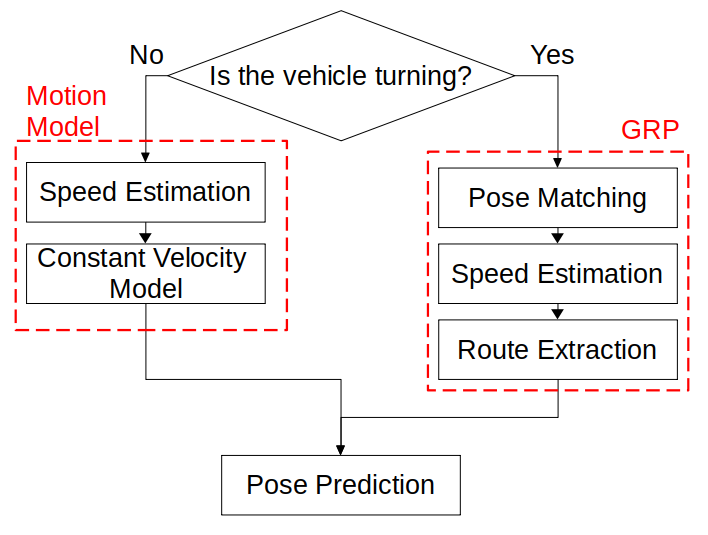}
    \caption{Overview of how a localization estimate is performed using the GRP}
\label{fig:localization estimate}
\end{figure}

The localization estimate of the next few frames is performed using the GRP and a motion model whenever a keyframe is selected. The keyframe will be fed into the bundle adjustment segment and the optimized poses of the previous few frames are used in the localization estimate. Within the estimation process, we measure the curvature of the past few poses to determine if the vehicle is turning. If the vehicle is going straight, the motion model is used and if it is turning, the GRP is used to predict the poses of the next few frames. This prediction is then saved for the next time a keyframe is selected, to be used as an initial estimate of the current pose in the bundle adjustment optimization. The process then repeats itself for every new keyframe selected. An overview of this localization estimate is shown in Fig. \ref{fig:localization estimate}.

\subsubsection{Motion Model}
In order to provide an estimate of the next frame's pose, ORB-SLAM uses a constant velocity motion model \cite{mur-artal_orb-slam_2015}. When testing this motion model prediction with DROID-SLAM, we found that we were able to approximate the pose of the vehicle accurately when it was travelling in a straight line but failed to model the pose of the vehicle during turns. This is understandable as the constant velocity model does not have information on the intent of the vehicle or the angle of turn. We address this limitation of the motion model at the turns using the global reference path.

\subsubsection{Global Reference Path}

We use the Open Source Routing Machine which is based upon the roads of OpenStreetMap \cite{haklay2008openstreetmap} to obtain a global reference path (GRP) for each unique route used for evaluation. The generated GRP was modified to match the driven path taken in the dataset. This route is then processed to have a higher fidelity and smoother turns by performing both interpolation and moving average smoothing. Orientation information is added by calculating the heading angle between each consecutive pair of points and the route is then aligned to the coordinate frame of the localization algorithm at the start point. Lastly, the scaling information is also required for alignment if the localization algorithm is unable to infer the scale. 

To find the ego-vehicle's position on the GRP, we adopt a matching procedure which searches for the point that best aligns with the current keyframe pose. 
This is done by first filtering all poses in the global route to be within a $1$ degree heading angle difference from the current pose. Then, the pose with the shortest Euclidean distance from the current pose is taken as the match. Using a constant acceleration motion model, we estimate the speed of the vehicle. With the speed and the time taken between each frame, we obtain an estimated distance that the vehicle travels for the duration of each frame. This is then used to extract the route along the global reference path starting from the matched pose, to be used as the estimated location of the vehicle for the next few frames.
The Menger Curvature constant is used to identify a turn. If the value exceeds $0.05$, the GRP is used for the localization estimate, otherwise the motion model is used. This combined localization estimate gives a more accurate initial estimate of the next keyframe pose. In contrast, the original DROID-SLAM algorithm initializes the next keyframe pose using the current frame's pose for bundle adjustment. 

\subsubsection{Conservative Global Reference Path (CGRP)}
A potential problem with our approach is that if the estimated keyframe pose is inaccurate, it might cause a worsening of the localization result. This inaccurate localization result consequently affects the accuracy of the next estimated keyframe pose, resulting in a negative feedback loop. Thus, we propose using a conservative localization estimate where the distance of the estimation from the previous keyframe pose is reduced by half to ensure that the estimate does not stray too far off. This is useful in the presence of heavy rain where visual data is significantly compromised with depth information being inaccurate leading to incorrect localization.

\begin{table*}[h]
\caption{Absolute Trajectory Errors(ATE) across Oxford Robotcar dataset sequences. Sorted from left (lowest blur) to right (highest blur). MDS - Modified DROID-SLAM, GRP - Global Reference Path, H - Heuristics, CGRP - Conservative Global Reference Path. The percentage improvement refers to the improvement of MDS + CGRP + H compared with MDS}
\centering
\begin{tabularx}{\linewidth}{ |c | >{\centering\arraybackslash}X |> {\centering\arraybackslash}X >{\centering\arraybackslash}X >{\centering\arraybackslash}X >{\centering\arraybackslash}X| >{\centering\arraybackslash}X| >
{\centering\arraybackslash}X|}
 \hline
 ATE(m) & \multicolumn{5}{|c|}{Oxford Robotcar Dataset} & Singapore Dataset & \multicolumn{1}{c|}{\multirow{2}{*}{}}\\
 \cline{1-7}
 
 Methods & 2014-12-09-13-21-02 (Clear) & 2015-10-29-12-18-17 (Rain) & 2014-11-25-09-18-32 (Rain) & 2015-05-29-09-36-29 (Rain + Alt Route) & 2014-11-21-16-07-03 (Rain + Night) & Singapore (Heavy Rain) & Average Across Rain Sequences\\
 \hline
 ORB-SLAM3  & - & 247.53 & 120.77 & - & - & - & -\\
 \hline
 MDS & 28.16 & 100.99 & 56.99 & 15.36 & 58.26 & 50.46 & 56.41\\
 MDS + GRP + H & \textbf{15.27} & \textbf{14.00} & \textbf{14.91} & \textbf{9.79} & \textbf{5.29} & 165.27 & 41.85\\
 MDS + CGRP + H & 18.49 & 14.77 & 23.18 & 11.56 & 15.62 & \textbf{44.66} & 21.96\\
 Percentage Improvement & 34.32\% & 85.38\% & 59.33\% & 24.77\% & 73.18\% & 11.50\% & 50.83\%\\

 \hline
\end{tabularx}
\label{table:1}
\end{table*}

\subsection{Heuristics} \label{heuristics}
We propose two heuristics for: (a) checking infeasible localization output and (b) dynamically modifying keyframe selection criteria along with the keyframe culling criteria with respect to the confidence of the correspondences.

\subsubsection{Infeasible Localization Output}
According to Berjoza et. al., for a two-axle automobile, the maximum turning angle it can achieve is 40 degrees \cite{berjoza2008research}. Thus, we measure the turning angle of DROID-SLAM's localization output and replace it with our localization estimate if it exceeds 40 degrees. Then, we cull the previous keyframe to ensure that this erroneous localization does not occur again.

\subsubsection{Dynamic modification of keyframe selection criteria}
DROID-SLAM uses a visual motion threshold (keyframe filter threshold) to decide when a keyframe is selected. We propose to dynamically adjust the keyframe filter threshold with respect to the confidence of the correspondence matching. The purpose is to identify when the visual data is severely compromised and to take counter-measures to prevent localization failure. DROID-SLAM provides a confidence prediction where a higher confidence represents a more reliable correspondence matching. The frequency of keyframe selection should increase the lower the confidence of the current frame. This is to allow for a higher rate of optimization to reduce the errors resulting from the lowered confidence. A linear scaling of the keyframe filter threshold using the confidence prediction from DROID-SLAM was implemented where the upper and lower bound of the confidence values are set to $0.1$ and $0.04$, respectively. For the keyframe filter threshold, the upper and lower bounds were set to $2.4$ and $1.8$, respectively. Also, $2.4$ is the default keyframe filter threshold used by DROID-SLAM.

When the image is too noisy, a higher keyframe selection frequency would not reduce errors. An example would be a sudden exposure of light when making a turn, where there is extreme noise for a short duration of time. To account for such scenarios, we minimize the keyframe selection frequency when confidence falls below a threshold of $0.04$ to minimize the localization error incurred during that period. We also prevent the culling of keyframes during this period to ensure continuity of the keyframes. This is set back to normal when the confidence rises back to above $0.1$. The threshold values used are obtained empirically.
Lastly, we also minimize keyframe selection frequency when the vehicle is at rest and ensure that a keyframe is selected after a maximum of 0.25s has passed to maintain continuity between the keyframes.

\section{Experimental Results}
\subsection{Datasets}
Our approach is tested on two datasets, our internal dataset collected in Singapore and the Oxford Robotcar dataset \cite{maddern_1_2017}. The Oxford Robotcar dataset provides a wide range of rain weather conditions including rain at different time of day and different traffic conditions for the same route. It uses the Bumblebee XB3 Trinocular stereo camera, providing up to $24$ cm stereo baseline at $16$ fps. There are nine rain sequences in the Oxford Robotcar dataset, only four were chosen out of the nine as the other five were either missing or severely lacking ground truth poses. An additional clear weather sequence was also included in the evaluation where all sequences except 2015-05-29 follow the same route of length around $9$km. Out of the five, the 2015-05-29 sequence had misaligned ground truth nearing the end which we cut short during our pre-processing of the data, while the 2014-11-21 sequence had an incomplete ground truth as both GPS and GNSS+INS data were compromised. Thus the routes of the two sequences shown in Fig. \ref{fig:visualization of robotcar results} are shorter than the other three.

In order to evaluate the robustness of our approach in heavier rain, we used heavy rain dataset from Singapore. This dataset uses two monocular NIR cameras which were setup in a stereo format facing outward at the front of the vehicle together with a GNSS+IMU system. We use the cameras to perform localization while the GNSS+IMU system provides the ground truth.

\subsection{Quantification of rain distortion}
We validate the performance of our localization algorithm with respect to the amount of distortion caused by rain. To quantify this distortion, we used a blur index. We followed the algorithm described in \cite{hanghang_tong_blur_2004} which uses the Haar wavelet transform to measure blurriness, which uses the "BlurExtent" ratio described in their paper. The higher the ratio, the more blurry the image is. The blurriness of every frame is measured using this method and the average value is taken as the blur index of the sequence. Table \ref{table:ablation} shows the blur index measured for each sequence.

\subsection{Evaluation Metric}
We use the absolute trajectory error (ATE) \cite{sturm_benchmark_2012} where the output poses are aligned and scaled (7DOF) before being projected onto the 2D plane for evaluation. DROID-SLAM \cite{teed_droid-slam_2021} and ORB-SLAM3 \cite{campos_orb-slam3_2021} are chosen as the baseline for comparison as they are the state-of-the-art localization algorithms in adverse and clear weather, respectively.

\begin{figure}
    \setlength\tabcolsep{1pt}
    \settowidth\rotheadsize{MDS + CGRP + H}
    \setkeys{Gin}{width=\hsize}
    \begin{tabularx}{\columnwidth}{l >{\centering\arraybackslash}X >{\centering\arraybackslash}X}
     & MDS & MDS + CGRP + H\\
    \rothead{\centering 2014-12-09 (clear)} &   \includegraphics[valign=m]{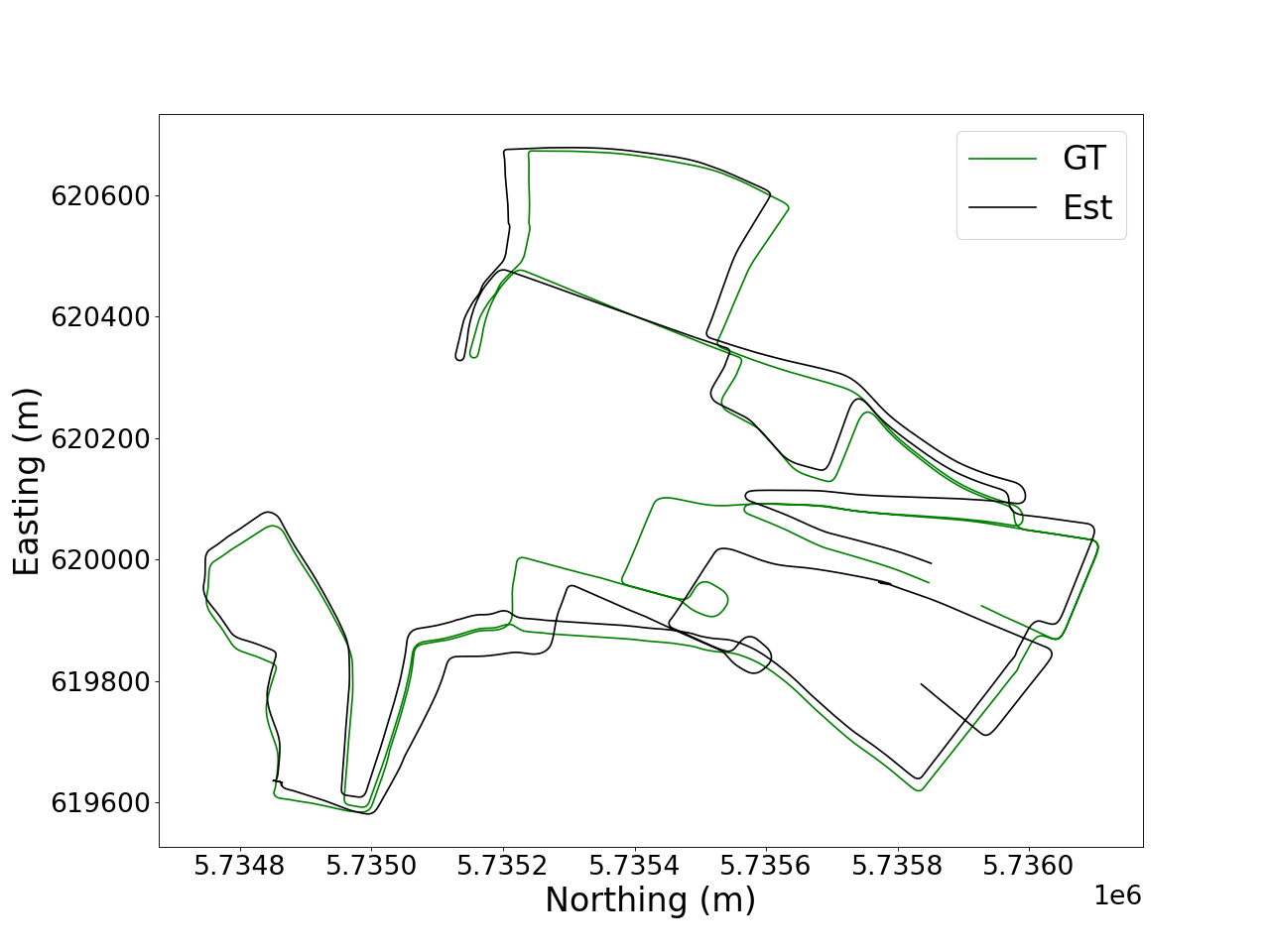}
                            &   \includegraphics[valign=m]{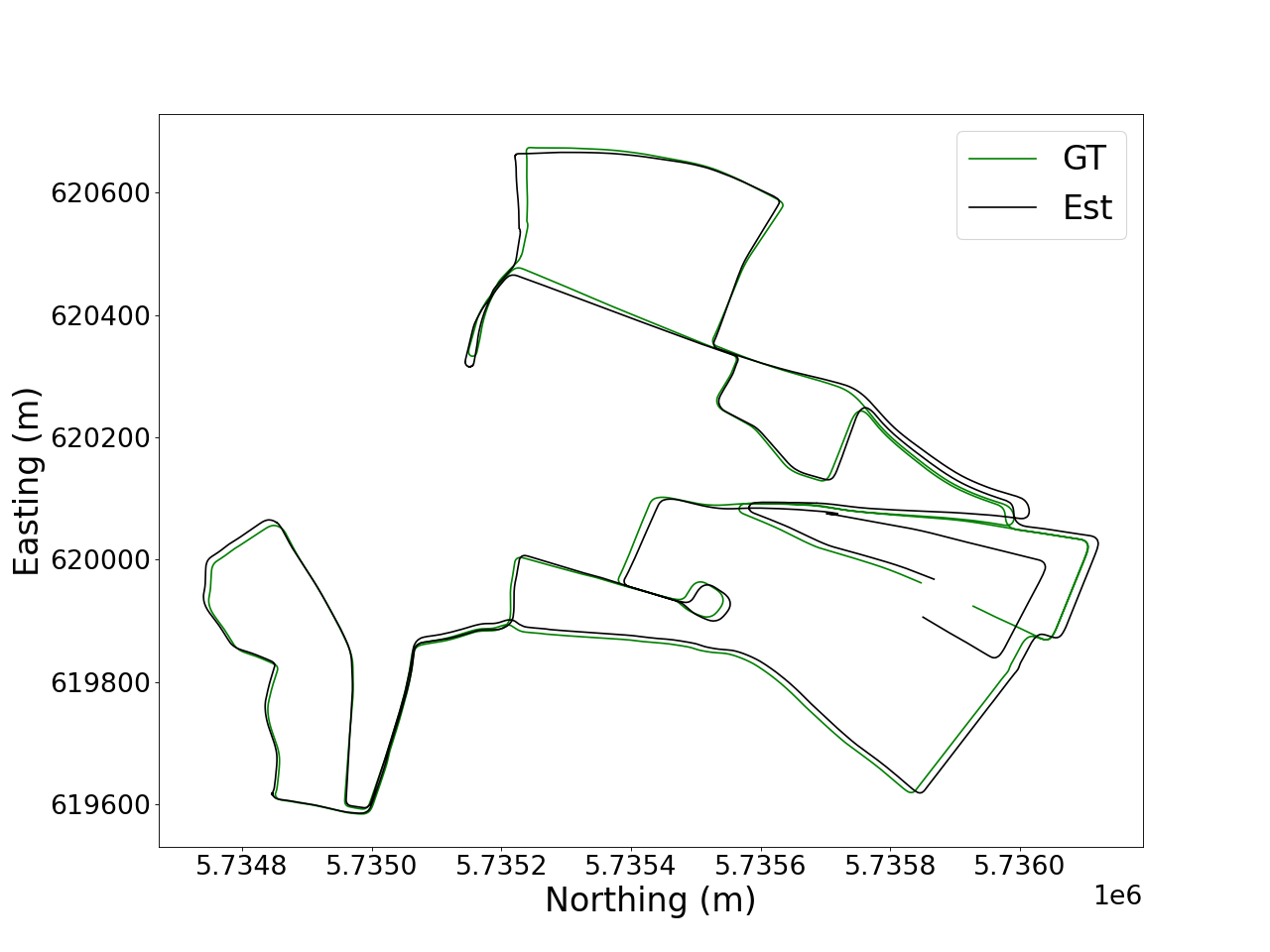}\\
    \rothead{\centering 2015-10-29 (rain)} &   \includegraphics[valign=m]{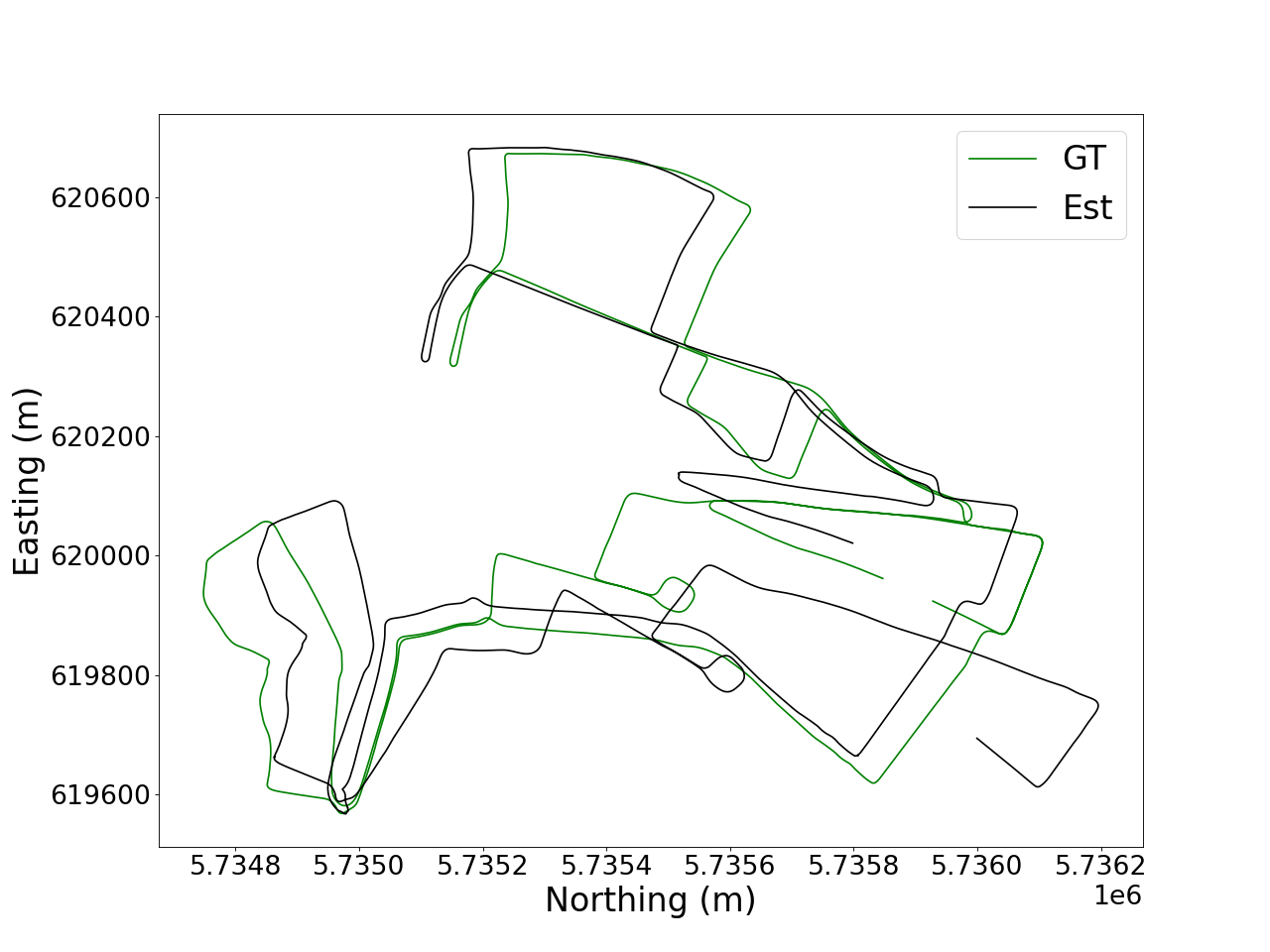}
                            &   \includegraphics[valign=m]{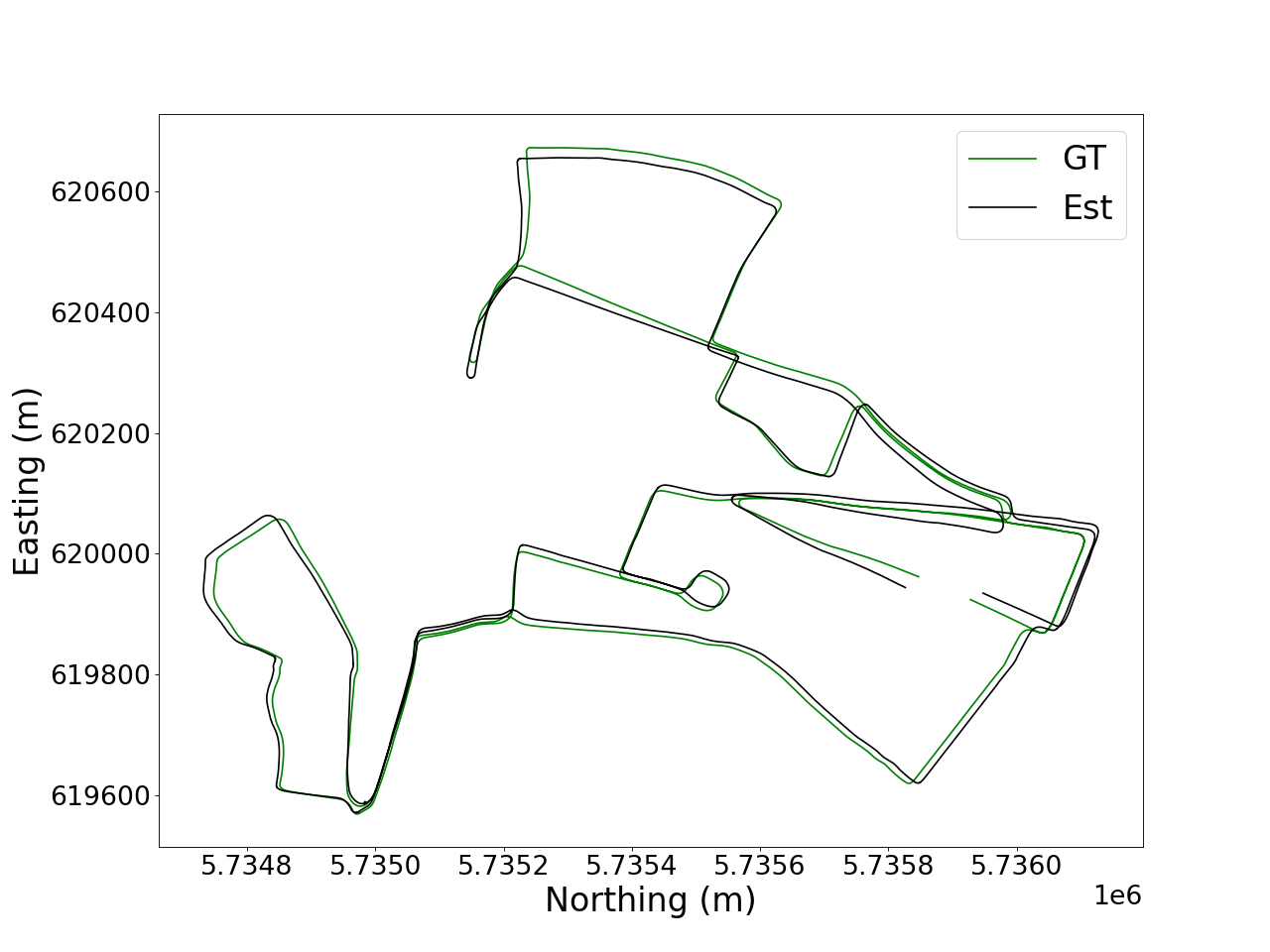}\\
    \rothead{\centering 2014-11-25 (rain)} &   \includegraphics[valign=m]{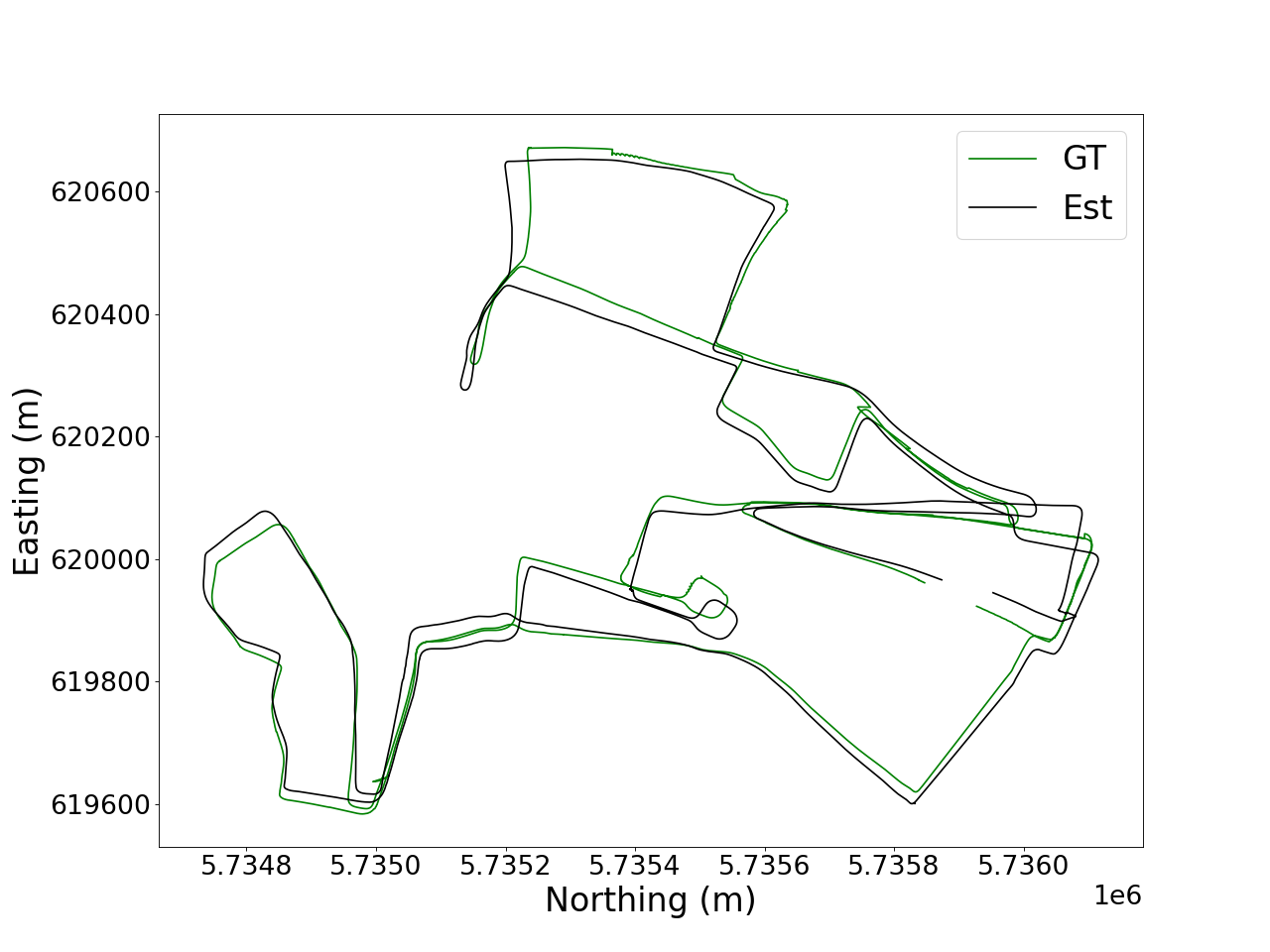}
                            &   \includegraphics[valign=m]{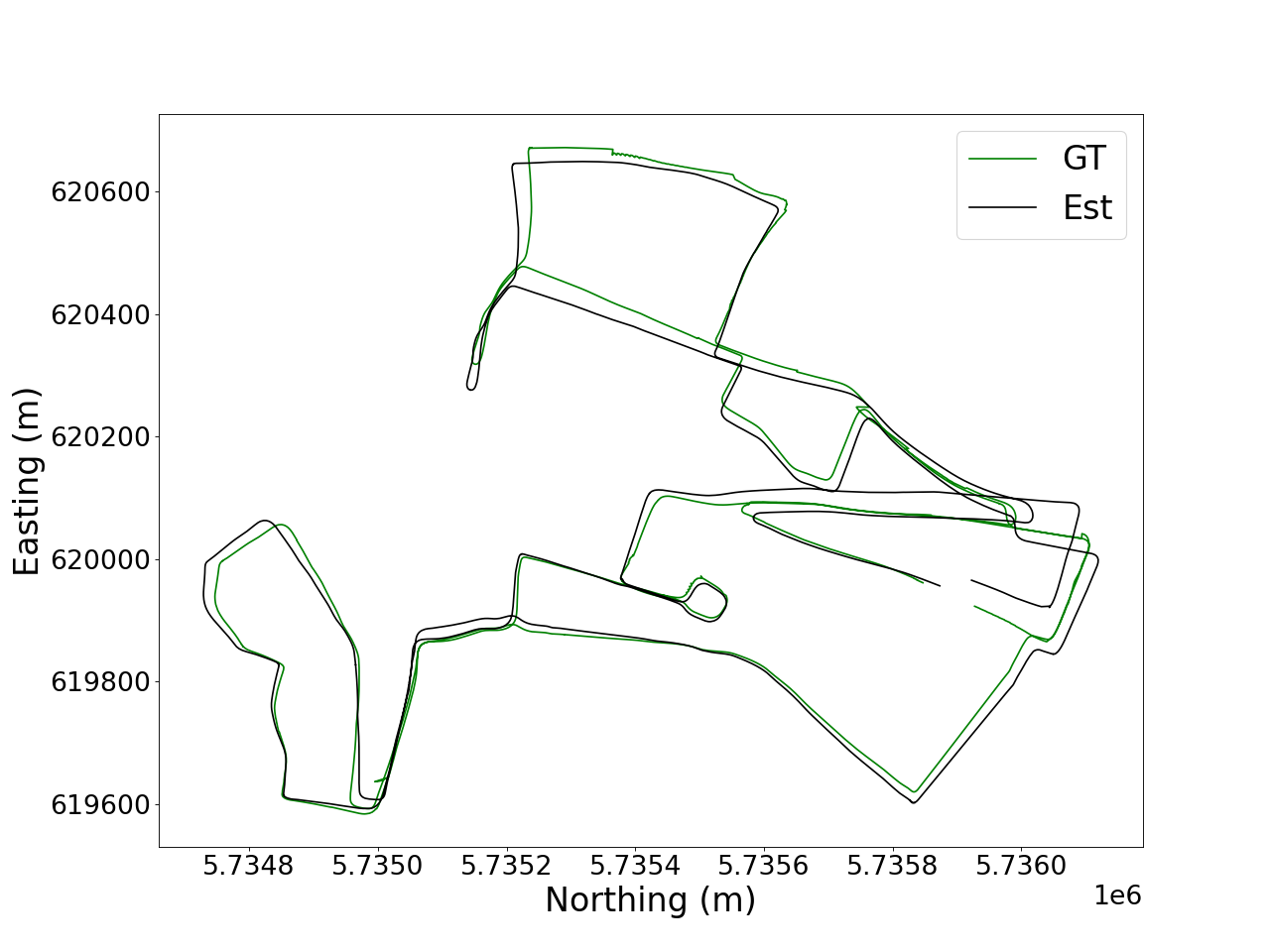}\\
    \rothead{\centering 2015-05-29 (rain + alt route)} &   \includegraphics[valign=m]{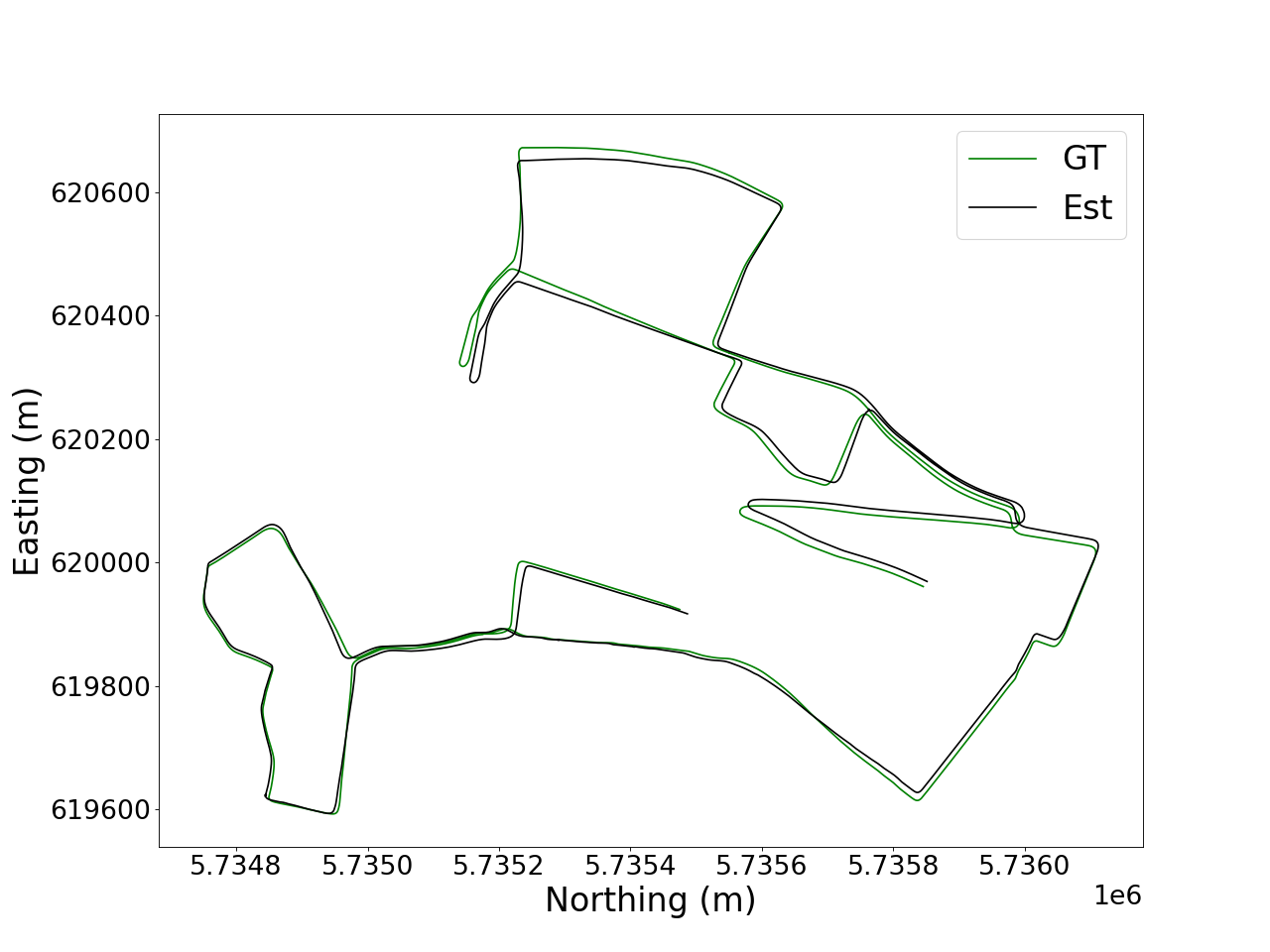}
                            &   \includegraphics[valign=m]{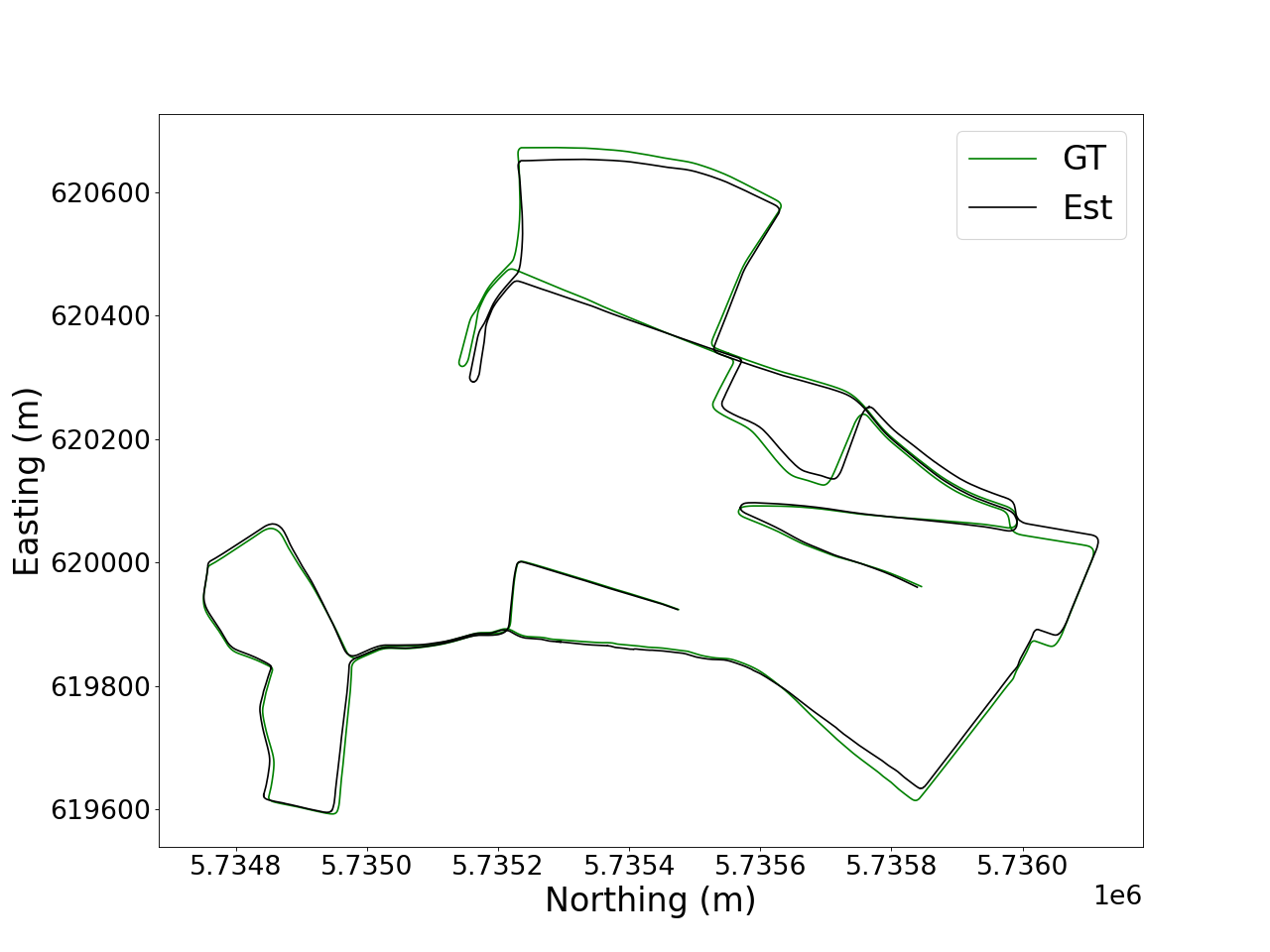}\\
    \rothead{\centering 2014-11-21 (rain + night))} &   \includegraphics[valign=m]{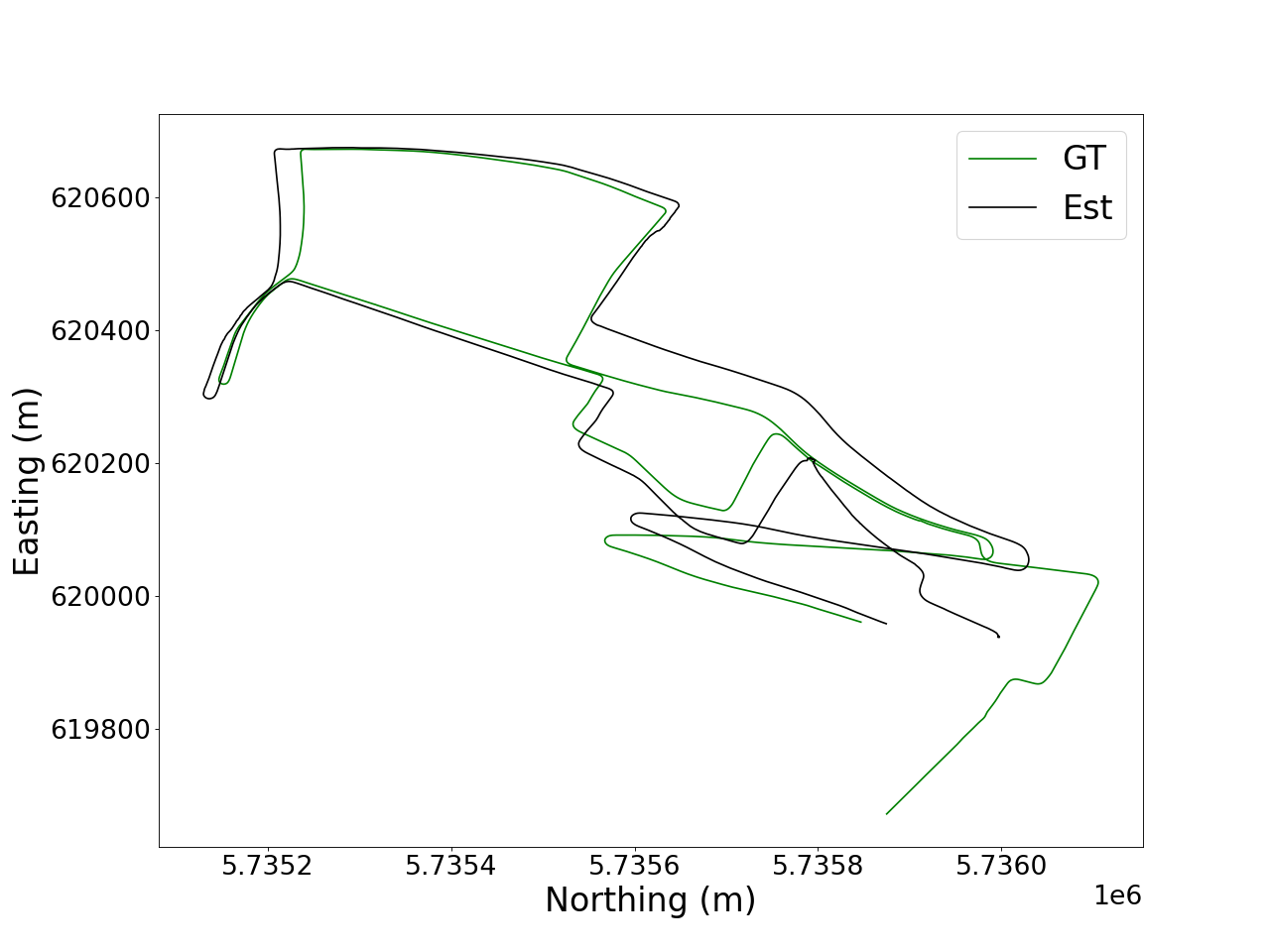}
                            &   \includegraphics[valign=m]{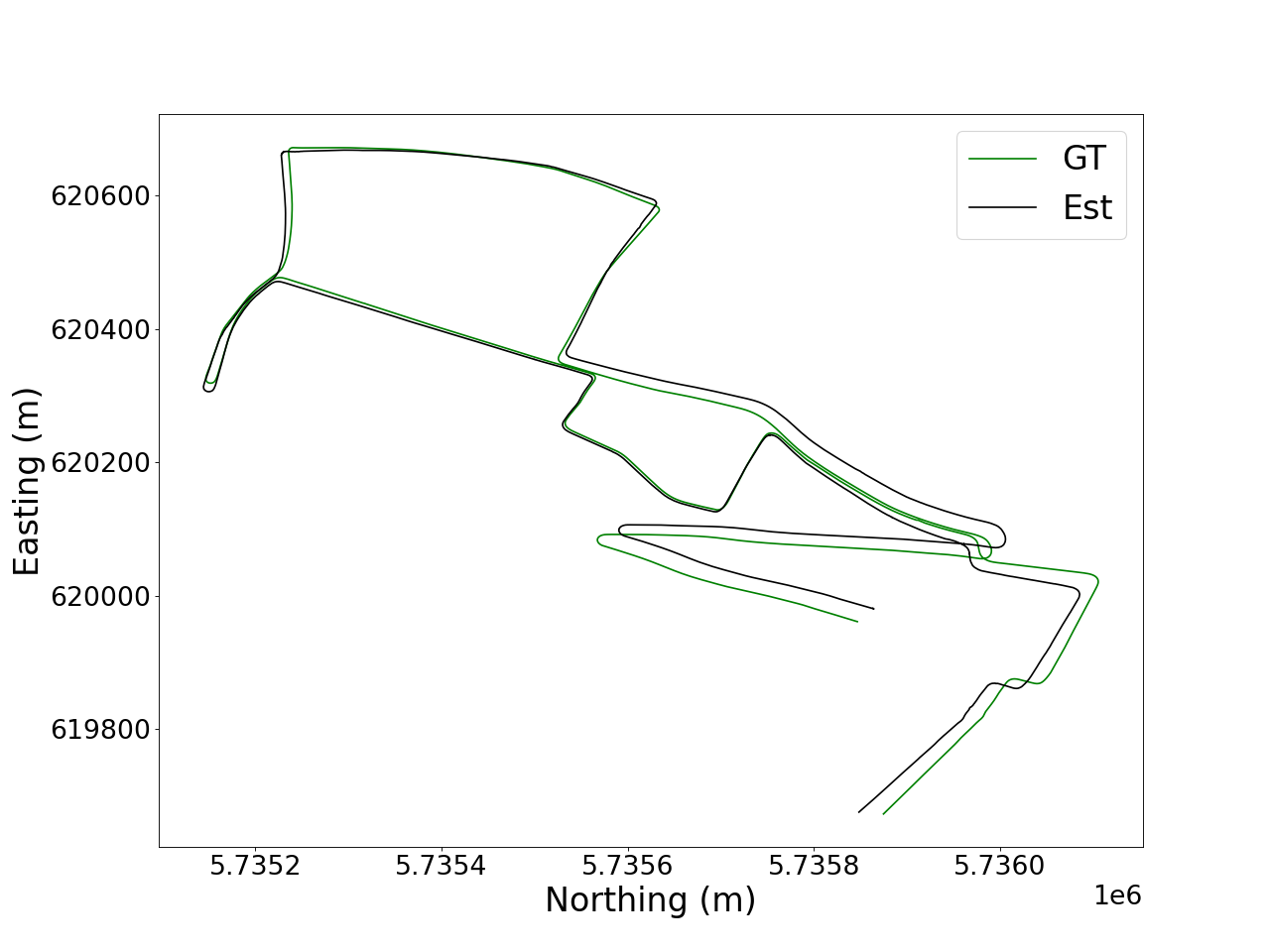}\\
    \rothead{\centering Singapore (heavy rain)} &   \includegraphics[valign=m]{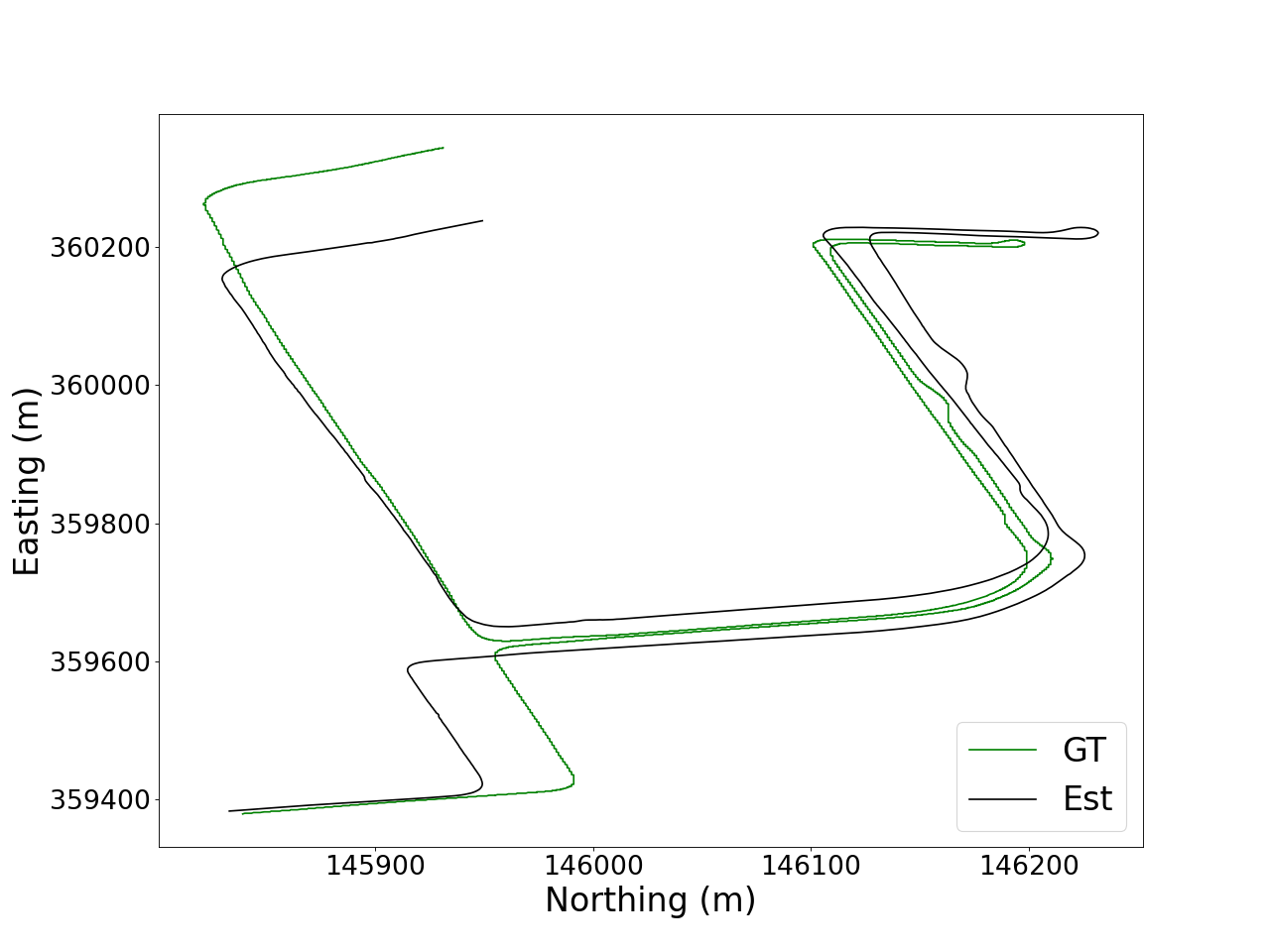}
                            &   \includegraphics[valign=m]{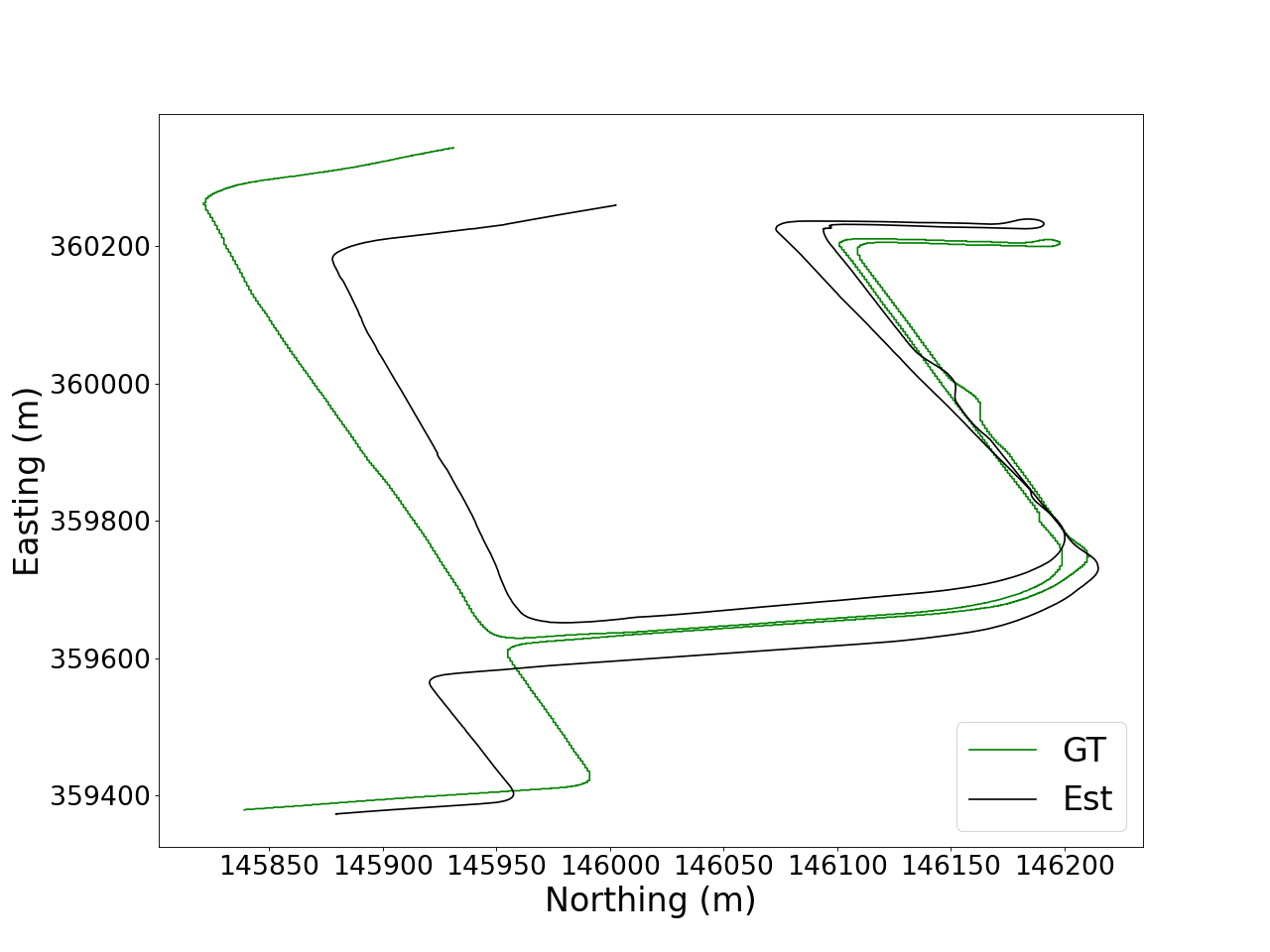}\\
    \end{tabularx}
    \caption{Visualization of the localization results before and after adding the conservative GRP and heuristic methods. The MDS result visualized here uses the 2.4 keyframe filter threshold which is the default value used by DROID-SLAM. MDS - Modified DROID-SLAM, CGRP - Conservative Global Reference Path, H - Heuristics}
    \label{fig:visualization of robotcar results}
\end{figure}

\subsection{Experimental Setup}
DROID-SLAM faces the limitation of large memory requirements as it saves every keyframe image and its corresponding features in order to perform global bundle adjustment at the end of the route. It runs out of memory quickly on a $6$GB GPU RAM when evaluating on the Oxford Robotcar dataset, resulting in an incomplete localization. In order to run DROID-SLAM for the entire route, we modified the algorithm to forget the past keyframes when it reaches a maximum of $6$GB memory capacity on the GPU. This disables the global bundle adjustment and the filling in of non-keyframe poses at the end of the route, thereby transforming the SLAM algorithm into a local visual odometry algorithm. The DROID-SLAM baseline that we compare against is this Modified DROID-SLAM (MDS).

For our proposed approaches, we setup the Modified DROID-SLAM with GRP and heuristics (MDS + GRP + H) meant for light rain scenarios, and the Modified DROID-SLAM with conservative GRP and heuristics (MDS + CGRP + H) meant for heavy rain scenarios. These two methods are compared with ORB-SLAM3 and the MDS for all sequences to evaluate the localization accuracy in different rain conditions.

For each sequence, three pre-processing steps were performed: (a) obtaining the global route from OpenStreetMap, (b) interpolating the ground truth such that a ground truth pose is generated at the timestamp for every image frame. The ground truth is also cleaned or trimmed if erroneous points are found and (c) a new start point is chosen where the vehicle is already on the main road to align the global reference path with the map coordinate frame. All experiments are conducted from this new start point.

\begin{table*}[t]
  \centering
  \caption{Ablation study of contributions across different Keyframe Filter Threshold. The sequences are sorted from top (lowest blur) to bottom (highest blur). MDS - Modified DROID-SLAM, GRP - Global Reference Path, H - Heuristics, CGRP - Conservative Global Reference Path}
    \begin{tabular}{|c|c|rrrr|r|c|}
    \hline
    \multicolumn{2}{|c|}{\multirow{2}[2]{*}{ATE (m) }} & \multicolumn{5}{c|}{\multirow{2}[1]{*}{Keyframe Filter Threshold }} & \multicolumn{1}{c|}{\multirow{2}[1]{*}{Blur Index }} \\
    \multicolumn{2}{|c|}{} & \multicolumn{5}{c|}{} & \multicolumn{1}{c|}{} \\
    \cline{1-7}
    \multicolumn{1}{|c|}{Sequences} & \multicolumn{1}{|c|}{Method} & \multicolumn{1}{c|}{1.2 } & \multicolumn{1}{c|}{1.8 } & \multicolumn{1}{c|}{2.4 } & \multicolumn{1}{c|}{3.0 } & \multicolumn{1}{c|}{Avg} & \multicolumn{1}{c|}{} \\
\hline
    \multicolumn{1}{|c|}{\multirow{4}[0]{*}{2014-12-09-13-21-02 (Clear)}} & MDS   & 25.89 & 16.48 & 51.77 & 18.49 & 28.16 & \multicolumn{1}{c|}{\multirow{4}[0]{*}{0.61}} \\
          & MDS + GRP & 22.30 & 16.13 & 40.58 & 32.98 & 28.00 & \\
          & MDS + GRP + H & \multicolumn{4}{c|}{15.27} & \textbf{15.27} & \\
          & MDS + CGRP + H & \multicolumn{4}{c|}{18.49} & 18.49 & \\
          \hline
    \multicolumn{1}{|c|}{\multirow{4}[0]{*}{2015-10-29-12-18-17 (Rain)}} & MDS   & 123.19 & 62.95 & 90.95 & 126.88 & 100.99 & \multicolumn{1}{c|}{\multirow{4}[0]{*}{0.72}}\\
          & MDS + GRP & 389.43 & 411.50 & 400.45 & 392.24 & 398.40 & \\
          & MDS + GRP + H & \multicolumn{4}{c|}{14.00} & \textbf{14.00} & \\
          & MDS + CGRP + H & \multicolumn{4}{c|}{14.77} & 14.77 & \\
          \hline
    \multicolumn{1}{|c|}{\multirow{4}[0]{*}{2014-11-25-09-18-32 (Rain)}} & MDS   & 45.27 & 77.71 & 25.20 & 79.77 & 56.99 & \multicolumn{1}{c|}{\multirow{4}[0]{*}{0.74}}\\
          & MDS + GRP & 50.32 & 19.80 & 25.58 & 26.09 & 30.45 & \\
          & MDS + GRP + H & \multicolumn{4}{c|}{14.91} & \textbf{14.91} & \\
          & MDS + CGRP + H & \multicolumn{4}{c|}{23.18} & 23.18 & \\
          \hline
    \multicolumn{1}{|c|}{\multirow{4}[0]{*}{2015-05-29-09-36-29 (Rain + Alt Route)}} & MDS   & 27.48 & 9.47 & 11.50 & 13.01 & 15.36 & \multicolumn{1}{c|}{\multirow{4}[0]{*}{0.75}}\\
          & MDS + GRP & 9.88 & 9.17 & 10.01 & 10.96 & 10.00 & \\
          & MDS + GRP + H & \multicolumn{4}{c|}{9.79} & \textbf{9.79} & \\
          & MDS + CGRP + H & \multicolumn{4}{c|}{11.56} & 11.56 & \\
          \hline
    \multicolumn{1}{|c|}{\multirow{4}[0]{*}{Singapore Route (Heavy Rain)}} & MDS   & 46.11 & 48.67 & 46.40 & 60.66 & 50.46 & \multicolumn{1}{c|}{\multirow{4}[1]{*}{0.79}}\\
          & MDS + GRP & 222.37 & 201.92 & 253.85 & 195.10 & 218.31 & \\
          & MDS + GRP + H & \multicolumn{4}{c|}{165.27} & 165.27 & \\
          & MDS + CGRP + H & \multicolumn{4}{c|}{44.66} & \textbf{44.66} & \\
          \hline
    \multicolumn{1}{|c|}{\multirow{4}[0]{*}{2014-11-21-16-07-03 (Rain + Night)}} & MDS   & 60.49 & 47.60 & 59.47 & 65.49 & 58.26 & \multicolumn{1}{c|}{\multirow{4}[0]{*}{0.80}}\\
          & MDS + GRP & 4.19 & 4.32 & 11.33 & 5.23 & 6.27 & \\
          & MDS + GRP + H & \multicolumn{4}{c|}{5.29} & \textbf{5.29} & \\
          & MDS + CGRP + H & \multicolumn{4}{c|}{15.62} & 15.62 & \\
          \hline
          
    \end{tabular}%
  \label{table:ablation}%
\end{table*}%

\subsection{Quantitative Evaluation}

Table \ref{table:1}, compares our method against the baseline algorithms across the sequences of varying rain conditions from both the Oxford Robotcar dataset and the Singapore dataset. We ran ORB-SLAM3 three times for every sequence and the best performing result is reported. ORB-SLAM3 is unable to complete the 2014-11-21 sequence and unable to start tracking for the Singapore route due to a lack of features. It also fails to track midway through the sequence for 2014-12-09 and 2014-05-29. Although ORB-SLAM3 is able to successfully track the entire sequence for 2015-10-29 and 2014-11-25, the localization result suffers greatly with the ATE being in the range of hundreds of meters. The ATE reported corresponds to the error of the longest path localized throughout the sequence. The Modified DROID-SLAM (MDS) is able to complete all routes successfully with the 2014-05-29 rain sequence outperforming the clear sequence. This might be due to the differing traffic conditions where 2014-05-29 had a less challenging traffic condition compared to the clear sequence. However, the average error across the rain sequences is higher than the clear sequence which signifies a reduction in localization accuracy in rain. With the introduction of the GRP and the heuristics, we are able to improve localization accuracy of the MDS for both clear and rain sequences except for the Singapore route while the CGRP and heuristics is able to improve localization accuracy for all sequences. The percentage improvement of MDS + CGRP + H over the MDS method shows that our contribution improved localization accuracy by $50.83$\% on average in rain sequences and $34.32$\% in the clear sequence. This gives evidence that the GRP together with our heuristic modifications is able to work in both clear and rain sequences, with greater improvements in rain conditions.

When analyzing Table \ref{table:1} from left to right, there is no clear pattern for the effect of visual quality on localization error. The inconsistency could be due to the varying traffic conditions and other confounding variables that changes between sequences. An average blur value of the route might also be insufficient where other methods of quantifying rain distortion such as the maximum blur within the route or the position of the blur within the image might give more insights to the effect of rain on localization accuracy. We note that ORB-SLAM3 is unable to perform localization once visual quality drops to the point where no matching features could be found while the MDS is still able to localize despite incurring some errors. Thus, a dense method such as DROID-SLAM would be more robust compared to a sparse method such as ORB-SLAM3 in rain weather conditions.

Fig. \ref{fig:visualization of robotcar results} shows the visualization of the localization results before and after the inclusion of our method. Qualitatively, for the MDS column, sequences 2014-12-09 and 2015-10-29 have high errors because of the wrongful localization at a particular segment of the route. Upon closer inspection of these segments, we found that the cause for such errors were due to either complex traffic conditions with large dynamic objects (buses or vans) or a short duration of over-exposure leading to tracking failure. While the errors for the 2014-11-21 sequences belong mostly to inaccurate localization of turns throughout the entire route due to the poor lighting and raindrop occlusions. Comparing between the two columns, our proposed methodology helps reduce the errors and provide a better alignment to the ground truth.

\vspace{-0.01\textheight}

\subsection{Ablation Study}
Table \ref{table:ablation} provides an ablation study of our contributions, where we incrementally validate the use of the GRP and the heuristics for improving localization accuracy. The GRP by itself provides minimal improvement to the MDS for the clear weather sequence but significantly improves the localization accuracy of MDS in rain weather, with the exception of the 2015-10-29 and the Singapore route sequence. Both MDS and MDS+GRP fails in the 2015-10-29 sequence due to a large dynamic object in a complex urban scenario. However, the GRP performs worse as the wrong localization result was matched in the GRP resulting in a wrong localization estimate. This estimate then further increases the error in the next localization iteration resulting in a negative feedback loop. In contrast, using our proposed heuristic, the negative feedback loop is avoided, lowering the localization ATE to $14.00$m.
In the Singapore route sequence, a negative feedback loop also occurred due to an inconsistent scaling issue for both the MDS + GRP and MDS + GRP + H method. After the first turn, the MDS wrongly localizes the scale of the scene resulting in an overestimation of the localized position which kick-starts the negative feedback loop. The negative feedback loop is avoided by using the conservative GRP, lowering the localization ATE to $44.66$m. Thus, we propose that the MDS + CGRP + H method be used for heavy rain scenarios.

Analyzing the results horizontally, across the keyframe filter threshold, MDS and MDS + GRP performs erratically depending on which keyframe filter threshold it uses. With the inclusion of the heuristics described in Section \ref{heuristics}, we are able to obtain results comparable to the minimum across the four keyframe filter threshold tested. We also show that our heuristics method gives a lower error compared to the average across the four threshold tested, thus showing the effectiveness of our contribution.

\vspace{-0.01\textheight}
\subsection{Computational Speed}
We executed all the algorithms on our system with a 3080ti GPU and an Intel i9 3.60GHz CPU. Our methods are implemented in python and we evaluated the additional time required to run the MDS + CGRP + H algorithm compared with the MDS algorithm across the Oxford Robotcar dataset. Using our method requires on average an additional $6.45$ms per frame of calculation time in clear weather conditions while it requires an additional $7.34$ms per frame in rain conditions averaging across the $4$ rain sequences from the Oxford Robotcar datasets. The GRP and heuristics method by itself do not require GPU to run.
\section{Conclusion}
Rain weather introduces noise in the form of raindrops and lens-flare to images. This occludes part of the image, resulting in a lost in visual features, negatively impacting localization accuracy. We propose to use a global reference path together with additional heuristic methods to provide a more accurate initial estimate to the bundle adjustment optimization thereby improving localization accuracy. We show that our contributions improve the state-of-the-art SLAM algorithm for both clear and rain weather conditions, validating the effectiveness of our method.

\section*{Acknowledgment}
This research is supported by the Ministry of Education, Singapore, under its Academic Research Fund Tier 2 MOE-T2EP50121-0022.

\bibliographystyle{unsrt}
\bibliography{IROS_reference_list}

\end{document}